\documentclass{article} 
\usepackage{iclr2027_conference,times}

\usepackage{amsmath,amsfonts,bm}

\def\eqref#1{equation~\ref{#1}}

\def\1{\bm{1}}

\DeclareMathAlphabet{\mathsfit}{\encodingdefault}{\sfdefault}{m}{sl}
\SetMathAlphabet{\mathsfit}{bold}{\encodingdefault}{\sfdefault}{bx}{n}

\usepackage{hyperref}       
\usepackage{url}            
\usepackage{booktabs}       
\usepackage{amsfonts}       
\usepackage{nicefrac}       
\usepackage{microtype}      
\usepackage{xcolor}         

\usepackage{graphicx} 
\usepackage{amsmath}
\usepackage{multirow}
\usepackage{wrapfig}
\usepackage{needspace}
\usepackage{placeins}
\usepackage{algorithm}
\usepackage{algpseudocode}
\usepackage{subcaption}
\usepackage{longtable} 
\usepackage{array}     

\title{EvoRubric: Self-Evolving Rubric-Driven RL for Open-Ended Generation}

\author{Xin Guan\thanks{This work was done during Xin Guan's internship at Alibaba Group.}, Xiaomeng Hu, Shen Huang\thanks{Corresponding authors.} \\
Zhenyi Wang, Bo Zhang, Zijian Li \\
Pengjun Xie, Bo Liu, Jiuxin Cao\footnotemark[2] \\
Alibaba Group
}

\iclrfinalcopy 
\begin{document}

\maketitle
\lhead{} 
\begin{abstract}
Reinforcement Learning (RL) has advanced Large Language Models (LLMs) in verifiable domains, while open-ended generation remains challenging due to the absence of definitive rewards. Rubric-based RL provides explicit evaluation criteria, but learning to construct these criteria remains challenging when final-answer correctness is not verifiable. We propose \textbf{EvoRubric}, a co-evolutionary RL framework that combines criterion-validity feedback, response discrimination, and peer agreement to learn rubrics for open-ended generation. A shared policy acts as both a Reasoner and a Rubric Generator, using its current responses and historical rubrics to discover new evaluation dimensions. To combine adaptive rubric discovery with a stable validity check, a frozen copy of the initial policy serves as the Meta-Verifier, while a frozen Grader scores responses against the retained criteria. Discriminative feedback, Leave-One-Out peer consensus, and a persistent memory pool transform this feedback into complementary rewards that jointly optimize both policy roles, closing the loop between response improvement and rubric discovery. EvoRubric improves over matched static and external evolving-rubric baselines across five benchmarks in Medical, Writing, and Science. Across three training seeds, it achieves five-benchmark averages of $56.28$ at 8B and $61.13$ at 14B, exceeding the strongest matched baselines by 3.09 and 2.06 points, respectively. Human audits assess criterion validity and response quality, and experiments with expert-initialized rubrics demonstrate compatibility with human priors.
\end{abstract}


\section{Introduction}
Reinforcement Learning (RL) has been widely adopted to enhance the capabilities of Large Language Models (LLMs). In particular, Reinforcement Learning with Verifiable Rewards (RLVR) has significantly elicited advanced reasoning and deliberation capabilities in objectively verifiable domains, such as math and coding, as evidenced by standout models like DeepSeek-R1 \citep{DBLP:journals/nature/GuoYZSWZXZMBZY025} and the OpenAI o-series \citep{DBLP:journals/corr/abs-2412-16720}. However, beyond these verifiable tasks, a vast majority of real-world queries are inherently open-ended and lack definitive, unique answers, making direct policy optimization profoundly difficult. To address this, recent studies \citep{RubricRewards, DBLP:journals/corr/abs-2508-12790, QuRL, DBLP:journals/corr/abs-2601-08430, DBLP:journals/corr/abs-2511-10507} have proposed Rubric-based RL methods. By delineating a series of explicit, verifiable criteria to assess various dimensions of a response, these methods offer a promising direction for optimizing models in open-ended scenarios.

Rubric-based optimization involves two complementary functions: constructing criteria that specify response quality and applying those criteria to produce reward signals. A Grader executes the resulting rubric, while the rubric determines which response properties receive supervision. As the policy improves, its responses expose new strengths and failure modes that the evaluation criteria should capture. This suggests treating the ability to construct useful rubrics as a learning objective alongside the ability to produce high-quality responses.

\begin{figure}[t]
    \centering
    \includegraphics[width=0.95\linewidth]{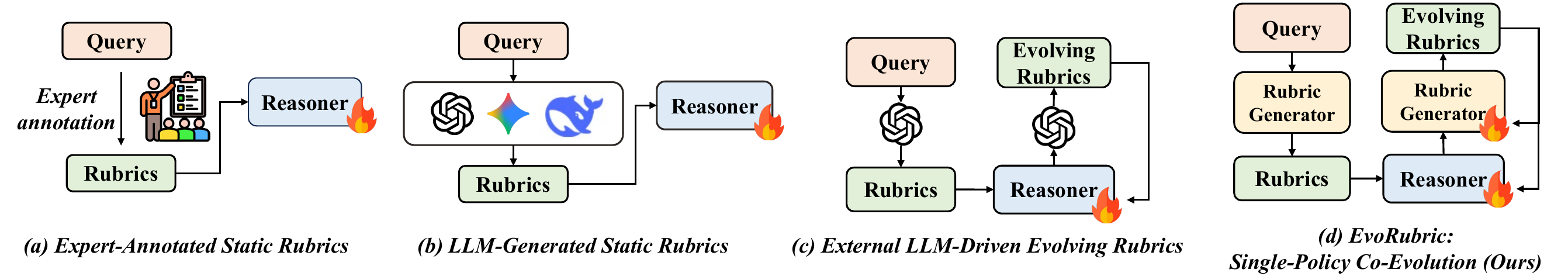}
    \caption{Rubric construction paradigms for open-ended generation: (a) expert-authored rubrics, (b) model-generated static rubrics, (c) external LLM-driven evolving rubrics, and (d) EvoRubric, which jointly optimizes response and rubric generation within a shared policy.}
    \label{fig:intro}
\end{figure}

Fixed rubrics can miss emerging response properties as the policy evolves (Figure~\ref{fig:intro}a,b). DR-Tulu \citep{DBLP:journals/corr/abs-2511-19399} addresses this through external rubric updates and a dynamic criterion pool (Figure~\ref{fig:intro}c). RLCER \citep{DBLP:journals/corr/abs-2602-10885} further trains a shared Reasoner and Rubricator, using correlation with final-answer correctness to supervise rubric quality. Open-ended generation often lacks this correctness reference. The resulting challenge is therefore \textit{how to learn useful evaluation criteria from current responses when final-answer correctness cannot provide the rubric-learning target}.

To address this challenge, we propose \textbf{EvoRubric}, which constructs rubric-learning feedback from criterion validity, response discrimination, and agreement among candidate rubric sets. A shared policy jointly learns response generation and rubric construction under these signals. The Reasoner samples candidate responses, whose differences provide concrete evidence for the Rubric Generator to discover new evaluation dimensions. Role-specific rewards train the Generator to produce valid, discriminative criteria and the Reasoner to improve its responses under these criteria. As the response distribution changes, the Generator learns to identify the evaluation dimensions that remain informative. We call this process \textit{self-evolving}: the policy learns to construct supervision from its own responses, uses that supervision to improve, and carries selected criteria forward through rubric memory. Rubric discovery becomes an integral part of policy learning without a separately parameterized rubric generator.

The feedback signals address complementary aspects of rubric quality. A frozen copy of the initial policy serves as the Meta-Verifier, screening criterion validity using the same model family and scale. A frozen Grader applies the retained criteria to responses. Discriminative feedback rewards criteria that separate current responses, LOO consensus measures agreement across rubric batches, and persistent memory carries selected criteria across updates. Together, these mechanisms provide rubric-learning supervision without a verifiable final-answer target.

We evaluate EvoRubric on five benchmarks across Medical, Writing, and Science. Across three training seeds, EvoRubric achieves the highest mean scores among matched methods on all five benchmarks at both 8B and 14B, improving the macro-average by 3.09 and 2.06 points, respectively. A stronger GPT-4.1-generator baseline, longitudinal rubric audits, and blind human response comparisons provide complementary evidence. Experiments with expert-initialized rubrics further show compatibility with human priors.

In summary, our main contributions are as follows:
\begin{itemize}
    \item We introduce \textbf{EvoRubric}, a rubric-learning framework for open-ended generation that constructs training feedback without a verifiable final-answer correctness reference.
    \item We integrate frozen initial-policy verification, discriminative feedback, and peer consensus with persistent rubric memory to support joint response and rubric learning.
    \item Independent-judge evaluation across three training seeds at both 8B and 14B demonstrates gains over matched baselines across five benchmarks. Human audits characterize residual errors and rubric diversity, and expert-initialization experiments show compatibility with human priors.
\end{itemize}


\section{Related Work}

\paragraph{Open-Ended QA.}
Driven by rapid advancements in Large Language Model (LLM) architectures and capabilities, alongside their significantly expanded context window capacities \citep{DBLP:conf/iclr/PengQFS24, DBLP:conf/acl/0001WY025, DBLP:journals/corr/abs-2601-10306}, LLM-generated responses are shifting from short-text tasks with deterministic answers (e.g., extractive QA \citep{DBLP:journals/tacl/TrivediBKS22} or multiple-choice questions \citep{rein2024gpqa}) to increasingly complex, open-ended long-text domains (e.g., medical question-answering \citep{DBLP:journals/corr/abs-2505-08775} and creative writing \citep{DBLP:journals/corr/abs-2503-05244}). Consequently, assessing the ``correctness'' and ``verifiability'' of these responses has become substantially more challenging. To address this, a growing body of research has pivoted towards the "LLM-as-a-Judge" paradigm \citep{DBLP:conf/acl/SongSSCM24,DBLP:journals/corr/abs-2506-11763}, leveraging LLMs to automatically evaluate response quality. More recently, to achieve more precise verification of response content and quality, emerging benchmarks have tended to design a series of fine-grained evaluation criteria to determine whether a response satisfies specific requirements. For instance, HealthBench \citep{DBLP:journals/corr/abs-2505-08775} incorporates expert-curated, medical-grade rubrics as checklists to thoroughly examine response quality. However, these annotation methodologies still heavily rely on human domain expertise or state-of-the-art closed-source models.

\paragraph{Reinforcement Learning with Rubrics.}
Reinforcement Learning with Verifiable Rewards (RLVR) has driven significant progress in deterministic domains like math and coding, empowering models such as DeepSeek-R1 \citep{DBLP:journals/nature/GuoYZSWZXZMBZY025} and the OpenAI o-series \citep{DBLP:journals/corr/abs-2412-16720}. However, extending this success to open-domain QA is challenging due to the difficulty of defining objective rewards for open-ended answers. Early RLHF methods measured human preferences via scalar scores \citep{DBLP:conf/nips/Ouyang0JAWMZASR22, DBLP:conf/nips/0001X024, DBLP:conf/iclr/DaiPSJXL0024}, but often lack interpretability and suffer from reward hacking and sparse feedback \citep{DBLP:journals/corr/abs-2502-18770, DBLP:journals/csur/ChaudhariAMRKNDS26}. Recently, Rubric-based RL has emerged, using high-quality rubrics as structured optimization signals \citep{DBLP:journals/corr/abs-2511-10507, DBLP:journals/corr/abs-2508-12790, DBLP:journals/corr/abs-2602-01511, DBLP:journals/corr/abs-2601-08430, QuRL, RubricRewards}. Rubric construction determines the supervision available to a policy. DR-Tulu \citep{DBLP:journals/corr/abs-2511-19399} updates externally generated criteria using current responses and a maintained pool. RLCER \citep{DBLP:journals/corr/abs-2602-10885} jointly optimizes a shared Reasoner and Rubricator with correctness-based rubric supervision. Rubric-ARM \citep{DBLP:journals/corr/abs-2602-01511} instead alternates optimization of a rubric generator and a judge against preference labels. EvoRubric jointly trains the response policy and rubric generator while keeping the Grader fixed, constructing rubric-learning feedback from validity checks and peer agreement when a verifiable correctness reference is unavailable. Appendix~\ref{sec:appendix_diff_previous_work} compares these supervision mechanisms.


\section{EvoRubric}
\paragraph{Problem formulation.}
We consider rubric-based RL for open-ended generation, where a rubric specifies evaluation criteria and a Grader applies them to candidate responses. Our objective is to learn high-quality responses and informative evaluation criteria without a verifiable final-answer correctness target. A shared policy $\pi_\theta$ performs response generation and rubric generation under role-specific prompts. Its initial checkpoint $\pi_{\theta_0}$ serves as a frozen Meta-Verifier, while a frozen Grader produces criterion-level response scores. Complementary rewards update $\theta$ for both generation roles.

\paragraph{Reinforcement Learning with GRPO.}
We employ Group Relative Policy Optimization (GRPO) \citep{DBLP:journals/corr/abs-2402-03300}. For each query and role, advantages are computed by standardizing rewards within the corresponding rollout group: $\hat A_i=(r_i-\mu)/(\sigma+\varepsilon)$, where $\mu$ and $\sigma$ are the group reward mean and standard deviation, and $\varepsilon$ ensures numerical stability. Reasoner and Generator rollouts form separate advantage groups and are then combined for a shared policy update using the clipped GRPO objective (Appendix~\ref{sec:app_grpo}).

\paragraph{Rubric-based scoring.}
A rubric $\mathcal C=\{c_1,\dots,c_K\}$ contains positive requirements and negative constraints with signed weights $w_k$. The Grader determines whether each criterion is met, yielding $s_{i,k}=w_k\mathbb I[c_k\text{ is met by }o_i]$, where $\mathbb I$ is the indicator function. Base, historical, and current rubric sets use the same signed-score normalization, yielding scores in $[-1,1]$. These scores provide fixed base-rubric and evolving rubric-pool feedback (Appendix~\ref{sec:reward_details}).

\begin{figure}[t]
    \centering
    \includegraphics[width=0.95\linewidth]{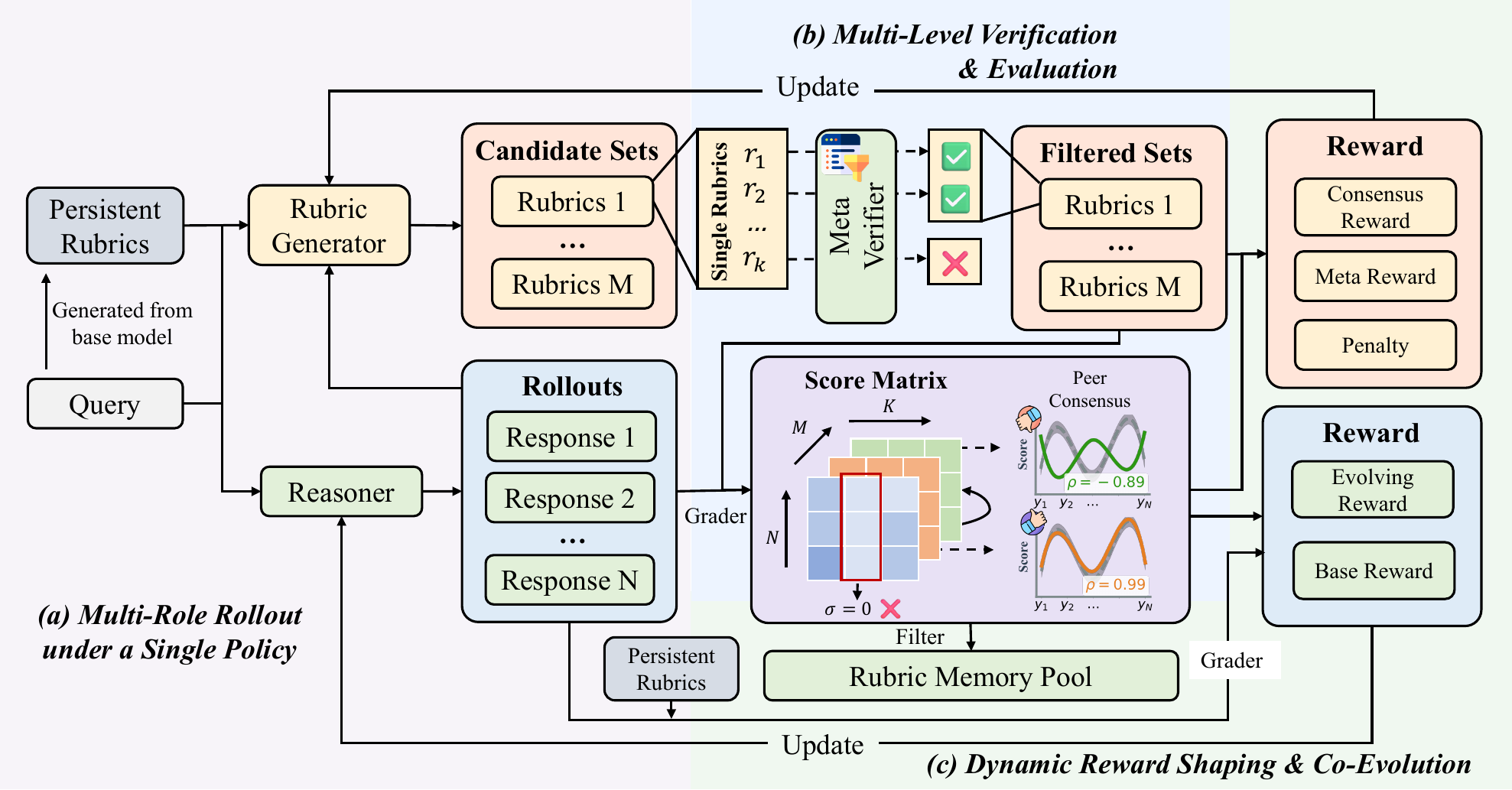}
    \caption{Overview of \textbf{EvoRubric}. A shared policy learns response generation and rubric discovery. A frozen initial-policy Meta-Verifier screens criteria, and a frozen Grader scores responses. Discriminative feedback, peer consensus, and rubric memory provide rewards for both policy roles.}
    \label{fig:overview}
\end{figure}

\subsection{Multi-Role Rollout under a Single Policy}
\label{sec:rollout}
EvoRubric couples response generation, criterion evaluation, and shared-policy optimization (Figure~\ref{fig:overview}). Let $p_{rea}$ and $p_{gen}$ denote the role-specific prompts for the Reasoner and Rubric Generator, respectively. Given a query $q$, the Reasoner first samples $N$ responses $\mathcal O=\{o_1,\dots,o_N\}$:
\begin{equation}
o_i\sim\pi_\theta(\cdot\mid q,p_{rea}).
\end{equation}
Within its input budget, the Generator then examines the sampled response text together with fixed base rubrics $\mathcal C_{base}$ and historical criteria $\mathcal C_{hist}$ retrieved from the Memory Pool, sampling $M$ candidate rubric sets:
\begin{equation}
R_m\sim\pi_\theta(\cdot\mid q,\mathcal O,\mathcal C_{base},\mathcal C_{hist},p_{gen}).
\end{equation}
Each set $R_m=\{r_{m,1},\dots,r_{m,K_m}\}$ targets evaluation dimensions missing from the existing criteria. Both roles use the current policy, so rubric discovery adapts to its evolving responses. We sample $N=8$ responses and $M=4$ rubric sets per query and maintain at most 10 evolved criteria in each query's pool. Fixed base rubrics are stored separately.

\subsection{Multi-Level Verification \& Evaluation}
\label{sec:verification}
\subsubsection{Motivation}
Rubric-based RL depends on criteria that provide useful reward differences among the policy's current responses. Without a verifiable final-answer target, the Generator needs feedback on both criterion validity and its value for learning. We combine a frozen initial-policy validity check with two complementary signals: \textbf{discriminative power (variance)}, which identifies criteria that separate current responses and can guide their relative improvement; and \textbf{peer consensus (correlation)}, which favors rubric batches whose relative scores agree on these responses. A valid criterion may become uninformative once all responses satisfy it, while a discriminative criterion may encode an unsupported requirement. Multi-level verification therefore evaluates validity, response separation, and cross-batch agreement together to train rubric discovery as the response distribution evolves.

\subsubsection{Verification signals}

\paragraph{Meta-verification and grading.}
The Meta-Verifier screens each criterion for factual or logical errors, unsupported requirements, and redundancy. With validity indicator $v(r)\in\{0,1\}$, the retained set is
\[
\tilde R_m=\{r\in R_m\mid v(r)=1\}.
\]
The Grader applies each retained criterion to all responses, producing the score vector $\mathbf s_r=[s(o_1,r),\dots,s(o_N,r)]^\top$.

\paragraph{Discriminative feedback.}
A criterion is discriminative when $\operatorname{std}(\mathbf s_r)>0$, separating responses within the current group. The proportion of such criteria contributes to the Generator reward. Historical criteria with zero variance are excluded from the current pool score, and saturated criteria for which all responses exhibit the desired behavior are removed as the pool evolves.

\paragraph{Leave-One-Out peer consensus.}
Let $F(o_i;\tilde R_m)$ denote the normalized score from a verified candidate batch and $\mathbf v_m=[F(o_1;\tilde R_m),\dots,F(o_N;\tilde R_m)]^\top$. For a batch with a nonconstant score vector and at least one nonconstant peer, we compute:
\[
\mathbf v_{\setminus m}=\frac{1}{|\mathcal P_m|}\sum_{j\in\mathcal P_m}\mathbf v_j,
\qquad \rho_m=\operatorname{corr}(\mathbf v_m,\mathbf v_{\setminus m}),
\]
where $\mathcal P_m$ contains the other nonconstant batches evaluated on the same responses. Consensus favors scoring patterns supported by multiple rubric groups, providing feedback for the current policy update alongside criterion-level verification. Its positive part is scaled by the fraction of proposed criteria passing meta-verification, yielding $\tilde\rho_m$. Degenerate cases and eligibility checks appear in Appendix~\ref{sec:reward_details}.

\subsection{Dynamic Reward Shaping \& Co-Evolution}
\label{sec:reward_coevolution}
\paragraph{Rubric Generator reward.}
The core reward combines criterion validity, discriminative value, and peer agreement:
\begin{equation}
R_{gen}(R_m)=\underbrace{\tilde\rho_m+r_{meta}+r_{disc}}_{\text{Quality \& Alignment}}
+\underbrace{(-p_{num}-p_{len})}_{\text{Regularization}}.
\end{equation}
Here, $r_{meta}$ is the fraction of proposed criteria passing meta-verification, and $r_{disc}$ rewards criteria that distinguish responses, scaled by this validity fraction. The penalties $p_{num}$ and $p_{len}$ discourage excessive criteria and overly long rubric text. Appendix~\ref{sec:reward_details} specifies batch eligibility and auxiliary regularization.

\paragraph{Rubric memory.}
Among eligible candidate batches, only the one with the highest LOO consensus contributes new criteria to the Memory Pool. Saturated historical criteria are removed, new entries are checked against the retained pool, and the oldest entries are evicted when capacity is exceeded. This selection carries useful evaluation dimensions into later rollouts while leaving the base rubrics fixed (Appendix~\ref{sec:memory_details}).

Memory connects criterion discovery across successive visits to a query, with stored criteria re-evaluated on each new response group. Retained requirements guide the Generator toward dimensions beyond the base rubrics and earlier discoveries. A criterion is saturated when every response exhibits the desired behavior: all satisfy a positive requirement or all avoid a penalized error. Such criteria can be retired to make room for new distinctions. Memory is maintained separately for each query, preserving the connection between a criterion and the task that motivated it.

\paragraph{Reasoner reward.}
The Reasoner is optimized against fixed base rubrics and evolving criteria that capture additional response-quality dimensions:
\begin{equation}
R_{rea}(o_i)=S_{base}(o_i)+S_{pool}(o_i)-p_{len}(o_i).
\end{equation}
Here, $S_{base}$ scores the response against the persistent base rubrics. $S_{pool}$ combines the historical score and the mean eligible current-batch score, each with weight $0.5$. Newly discovered criteria guide the current update, while the Memory Pool carries selected criteria into subsequent rollouts. The length penalty $p_{len}(o_i)$ increases linearly once the generated output exceeds a token-length threshold (Appendix~\ref{sec:reward_details}).

\paragraph{Joint update.}
Advantages are computed separately for the $N$ Reasoner samples and $M$ Generator samples of each query, and the combined rollout batch updates the shared policy. This produces a \textit{self-evolving} loop at two levels: policy updates improve response generation and rubric construction, while memory updates retain selected criteria and retire saturated ones. The next rollout uses both the updated policy and the revised memory. Responses thus supply evidence for learning new criteria, and those criteria supply supervision for improving responses, with the base rubrics, Meta-Verifier, and Grader held fixed.

The full training procedure is given in Appendix~\ref{sec:app_grpo} (Algorithm~\ref{alg:evorubric}).


\section{Experiments}
\label{sec:experiments}

\subsection{Experimental Setup}
\label{sec:experimental_setup}
\paragraph{Models and training configuration.}
We instantiate the shared policy with Qwen3-8B and Qwen3-14B \citep{DBLP:journals/corr/abs-2505-09388}. For each backbone, the Meta-Verifier uses the corresponding frozen pre-RL checkpoint. GPT-OSS-120B \citep{DBLP:journals/corr/abs-2508-10925} serves as the frozen training Grader. Training settings are summarized in Appendix~\ref{sec:app_hyperparameters}.

\paragraph{Datasets and benchmarks.}
We train on three domains, sampling 3,000 queries each from HealthBench (Medical) \citep{DBLP:journals/corr/abs-2505-08775}, Long-writer-zero (Writing) \citep{DBLP:journals/corr/abs-2506-18841}, and ResearchQA (Science) \citep{DBLP:journals/corr/abs-2509-00496}. For each backbone, we train a separate policy in each domain and evaluate it on the corresponding benchmarks. Evaluation covers a held-out set of 500 HealthBench instances, LLMEval-Med \citep{DBLP:conf/emnlp/ZhangSLSHWHLTJCXDGZH25}, WritingBench \citep{DBLP:journals/corr/abs-2503-05244}, Creative Writing \citep{creative-writing-bench-v3}, and held-out ResearchQA queries.

\paragraph{Matched baselines.}
Static Rubric-RL uses fixed rubrics generated before RL by the corresponding frozen base model with few-shot prompting. External Evolving-RL uses a separate, unoptimized same-scale rubric generator. These controlled comparisons share the policy backbone, training data, and policy-side configuration. Following the external rubric-generation strategy of DR-Tulu \citep{DBLP:journals/corr/abs-2511-19399}, we also evaluate a GPT-4.1-generator variant in Medical.

\paragraph{Evaluation protocol.}
We follow each benchmark's official evaluation scripts and scoring protocol, retaining its evaluation criteria while replacing the model judge. Unless otherwise specified, evaluations use \texttt{claude-sonnet-4-6} with default settings. This third-party evaluator belongs to a different model family from both the GPT-OSS training Grader and the Qwen3 policy and Meta-Verifier, reducing direct coupling between training-time scoring and final evaluation.

\begin{table}[t]
\centering
\setlength{\tabcolsep}{3pt}
\caption{
Main results (mean $\pm$ standard deviation over three training seeds for RL methods or three response-sampling runs for Base). Avg is the unweighted five-benchmark mean. Relative gains use the strongest baseline in each column within the same backbone; bold marks the best result.
}
\label{tab:main_results}
\resizebox{\linewidth}{!}{
\begin{tabular}{lcccccc}
\toprule
\multirow{2}{*}{\textbf{Method}}
& \multicolumn{2}{c}{\textbf{Medical}}
& \multicolumn{2}{c}{\textbf{Writing}}
& \multicolumn{1}{c}{\textbf{Science}}
& \multirow{2}{*}{\textbf{Avg}} \\
\cmidrule(lr){2-3}\cmidrule(lr){4-5}\cmidrule(lr){6-6}

& HealthBench 
& LLMEval-Med 
& WritingBench 
& Creative Writing 
& ResearchQA 
& \\
\midrule

\multicolumn{7}{l}{\textit{Qwen3-8B}} \\

Base
& $34.19\pm0.18$
& $49.25\pm0.22$
& $60.99\pm0.16$
& $41.05\pm0.27$
& $56.04\pm0.20$
& $48.30$ \\

Static Rubric-RL
& $38.89\pm0.34$
& $52.95\pm0.47$
& $65.06\pm0.72$
& $41.78\pm1.08$
& $65.83\pm0.39$
& $52.90$ \\

External Evolving-RL
& $39.53\pm0.55$
& $51.73\pm0.61$
& $65.98\pm0.94$
& $42.36\pm1.27$
& $66.34\pm0.48$
& $53.19$ \\

\textbf{EvoRubric}
& $\mathbf{45.13\pm0.71}$
& $\mathbf{53.75\pm0.84}$
& $\mathbf{67.59\pm0.89}$
& $\mathbf{45.91\pm1.34}$
& $\mathbf{69.02\pm0.57}$
& $\mathbf{56.28}$ \\

\midrule
\textit{Relative gain (\%)}
& $\mathbf{14.2\uparrow}$
& $\mathbf{1.5\uparrow}$
& $\mathbf{2.4\uparrow}$
& $\mathbf{8.4\uparrow}$
& $\mathbf{4.0\uparrow}$
& $\mathbf{5.8\uparrow}$ \\

\midrule

\multicolumn{7}{l}{\textit{Qwen3-14B}} \\

Base
& $37.91\pm0.14$
& $55.86\pm0.19$
& $65.12\pm0.13$
& $52.22\pm0.24$
& $61.55\pm0.17$
& $54.53$ \\

Static Rubric-RL
& $44.36\pm0.21$
& $58.30\pm0.25$
& $67.13\pm0.58$
& $56.59\pm0.90$
& $68.86\pm0.43$
& $59.05$ \\

External Evolving-RL
& $44.37\pm0.62$
& $58.02\pm0.32$
& $67.40\pm1.17$
& $56.38\pm1.35$
& $69.18\pm0.25$
& $59.07$ \\

\textbf{EvoRubric}
& $\mathbf{46.92\pm0.83}$
& $\mathbf{59.23\pm0.91}$
& $\mathbf{69.73\pm1.03}$
& $\mathbf{58.88\pm1.52}$
& $\mathbf{70.88\pm0.52}$
& $\mathbf{61.13}$ \\

\midrule
\textit{Relative gain (\%)}
& $\mathbf{5.7\uparrow}$
& $\mathbf{1.6\uparrow}$
& $\mathbf{3.5\uparrow}$
& $\mathbf{4.0\uparrow}$
& $\mathbf{2.5\uparrow}$
& $\mathbf{3.5\uparrow}$ \\

\bottomrule
\end{tabular}}
\end{table}

\subsection{Overall Performance}
\label{sec:main_results}
EvoRubric achieves the highest mean scores among matched methods on all five benchmarks at both scales (Table~\ref{tab:main_results}). Its macro-average scores of 56.28 and 61.13 exceed the strongest matched baselines by 3.09 and 2.06 points, corresponding to relative gains of 5.8\% and 3.5\%. At 8B, the largest relative improvements occur on HealthBench (14.2\%) and Creative Writing (8.4\%). Gains also extend to Chinese LLMEval-Med, where EvoRubric improves by 0.80 and 0.93 points at 8B and 14B. Together, these results demonstrate consistent mean improvements across Medical, Writing, and Science under an independent evaluator.

\subsection{Component and Generator Analysis}
\label{sec:ablation}

Table~\ref{tab:ablation} examines individual training components. The w/o Generator Loss variant updates the shared policy using only the Reasoner loss, retaining rubric generation, verification, grading, memory, and all selection rules. Rubric generation can therefore still change indirectly through shared parameters. Other variants remove verifier screening and feedback, historical memory, peer consensus, or discriminative feedback; the memory ablation retains current-rollout supervision.

\begin{wraptable}{R}{0.6\textwidth}
  \centering
  \setlength{\tabcolsep}{3pt}
  \caption{Medical component ablations (Qwen3-8B), averaged over 3 matched runs.}
  \label{tab:ablation}
  \resizebox{\linewidth}{!}{
  \begin{tabular}{l cc}
  \toprule
  \textbf{Model Variant} & \textbf{HealthBench} & \textbf{LLMEval-Med} \\
  \midrule
  w/o Generator Loss                  & 41.58 & 49.93 \\
  w/o Discriminative Feedback           & 38.92 & 51.11 \\
  w/o Peer Consensus                  & 40.80 & 50.16 \\
  w/o Memory Pool                     & 42.26 & 52.41 \\
  w/o Meta-Verifier                   & 43.13 & 47.81 \\
  \midrule
  \textbf{EvoRubric (Full)}           & \textbf{45.13} & \textbf{53.75} \\
  \bottomrule
  \end{tabular}
  }
\end{wraptable}

Removing the Generator loss lowers HealthBench and LLMEval-Med by 3.55 and 3.82 points, respectively. This supports the contribution of explicit rubric learning beyond the indirect changes induced by Reasoner updates. Every other ablation also lowers both scores: discriminative feedback has the largest effect on HealthBench, while removing meta-verification reduces LLMEval-Med to 47.81, below the base model's 49.25. The final-pool audit (Table~\ref{tab:human_errors}) links the latter decline to a higher incidence of invalid criteria. Removing peer consensus and memory reduces HealthBench by 4.33 and 2.87 points, respectively. These components support different aspects of adaptive supervision: consensus favors agreement among rubric groups on the current responses, whereas memory carries selected criteria into later rollouts for re-evaluation. The two ablations connect these mechanisms to improvements in final benchmark performance as the response distribution changes. Together, these results support both direct Generator optimization and the complementary rubric-quality signals.

\paragraph{External generation following DR-Tulu.}
\label{sec:strong_generator}
We instantiate DR-Tulu's external rubric-generation strategy \citep{DBLP:journals/corr/abs-2511-19399} in our Medical setting by using \texttt{gpt-4.1-2025-04-14} to propose evolving criteria from current responses and existing rubrics. With the same Qwen3-8B Reasoner backbone, the GPT-4.1 generator improves HealthBench/LLMEval-Med by 2.72/1.19 points over the frozen same-scale generator (Table~\ref{tab:generator_comparison}). EvoRubric gains a further 2.88/0.83 points, showing that rubric construction learned within an 8B policy can provide more effective training supervision than this strong external generator. Generator usage and cost are reported in Appendix~\ref{sec:strong_generator_details}.

\begin{table}[htbp]
\centering
\caption{Rubric-generation strategies in Medical (Qwen3-8B; means over 3 runs). The GPT-4.1 variant adopts DR-Tulu's external evolving-rubric generation strategy within our training setting.}
\label{tab:generator_comparison}
\setlength{\tabcolsep}{5pt}
\begin{tabular}{llrr}
\toprule
Method & Rubric generator & HealthBench & LLMEval-Med \\
\midrule
External Evolving-RL & Frozen Qwen3-8B & 39.53 & 51.73 \\
External (DR-Tulu-style) & GPT-4.1 & 42.25 & 52.92 \\
\textbf{EvoRubric} & \textbf{Shared learned policy} & \textbf{45.13} & \textbf{53.75} \\
\bottomrule
\end{tabular}
\end{table}

\subsection{Training Dynamics and Rubric Initialization}
\label{sec:additional_analysis}
\subsubsection{Training dynamics}
\begin{figure}[htbp]
  \centering
  \begin{minipage}{0.45\textwidth}
    \centering
    \includegraphics[width=\linewidth]{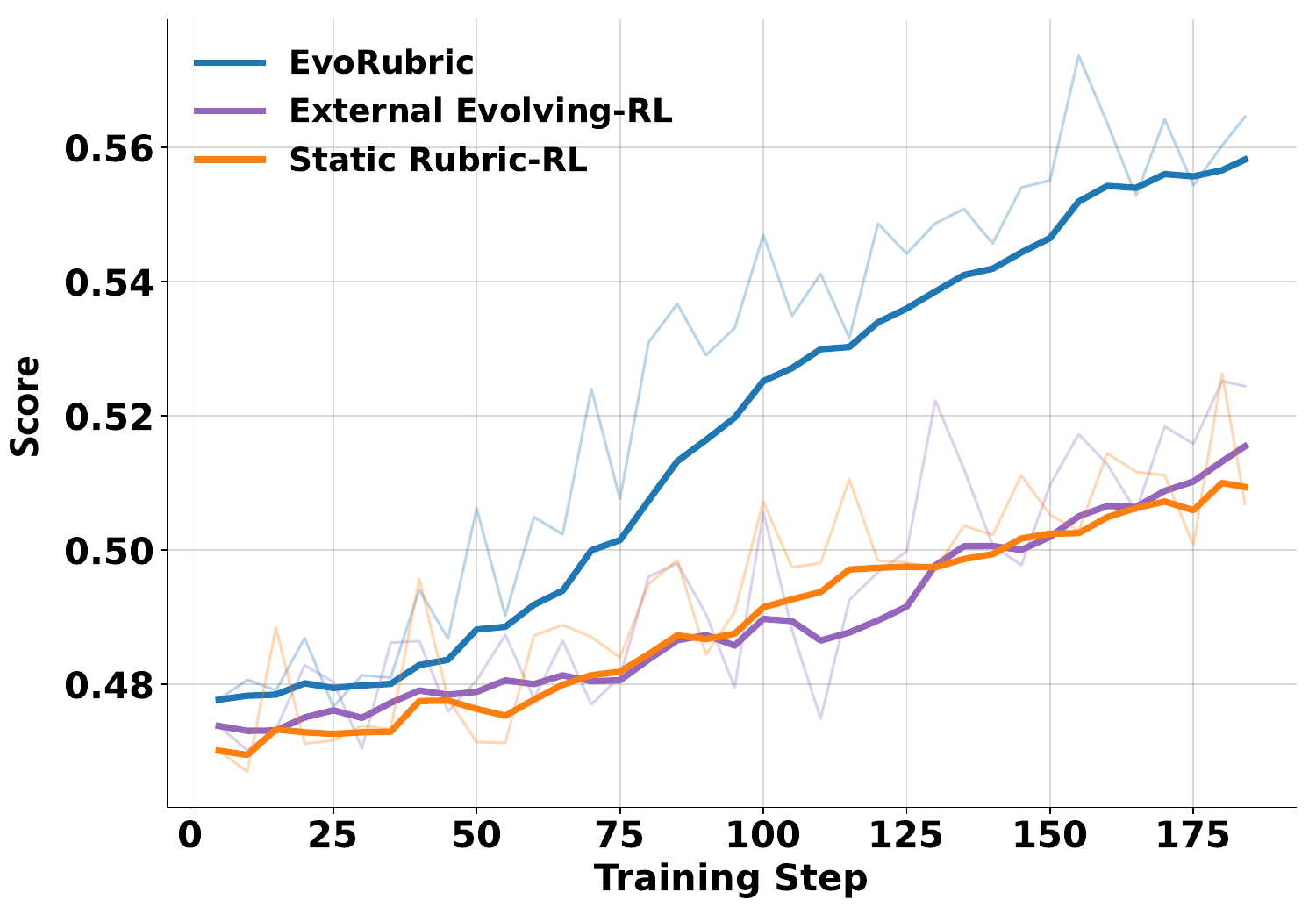}
    \subcaption{Standard training dynamics}
  \end{minipage}\hfill
  \begin{minipage}{0.45\textwidth}
    \centering
    \includegraphics[width=\linewidth]{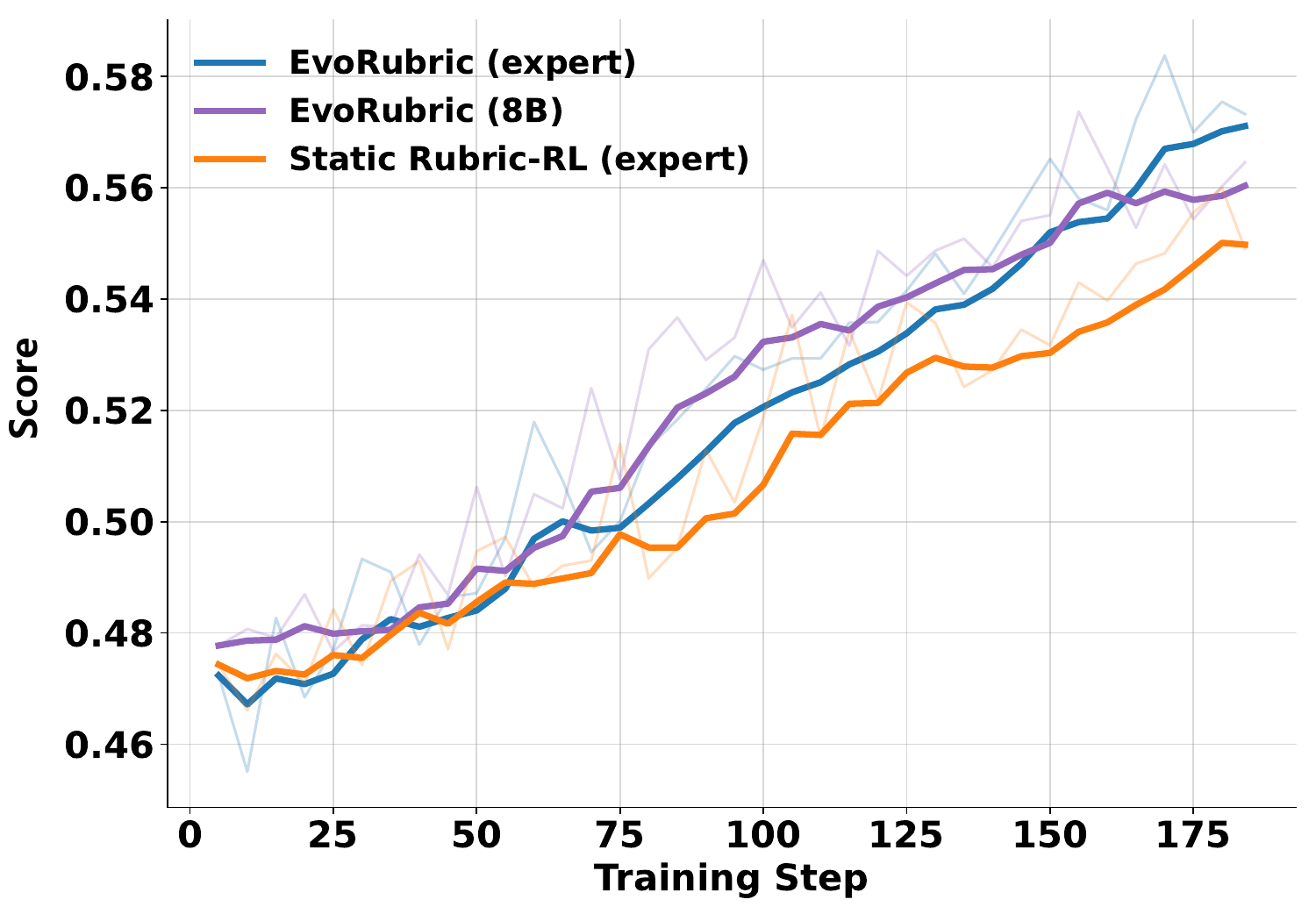}
    \subcaption{Impact of expert base rubrics}
  \end{minipage}
  \caption{Medical learning curves showing GPT-OSS-120B scores under base rubrics held fixed throughout RL. (a) Model-generated base rubrics. (b) Expert versus model initialization.}
  \label{fig:training_curves}
\end{figure}

Figure~\ref{fig:training_curves} shows that, with model-generated base rubrics, EvoRubric reaches higher scores than the static and external evolving-rubric baselines. With expert-initialized rubrics, co-evolution also improves the learning trajectory, motivating the initialization comparison below.

The trajectories begin at similar score levels and separate as training proceeds, with EvoRubric maintaining a larger advantage in the later stages. Because the base rubrics remain fixed within each initialization setting, these curves track progress against a stable evaluation target while the Memory Pool evolves. The held-out initialization results below examine whether this training-time advantage carries over to final response quality.

\FloatBarrier
\Needspace{4\baselineskip}
\subsubsection{Expert-initialized rubrics}
\begin{wraptable}{r}{0.56\textwidth}
  \centering
  \caption{Rubric initialization on HealthBench (Qwen3-8B; means over 3 runs).}
  \label{tab:expert_eval}
  \begin{tabular}{@{}p{\dimexpr\linewidth-6.5em-2\tabcolsep\relax}r@{}}
  \toprule
  \textbf{Method} & \textbf{HealthBench} \\
  \midrule
  Static Rubric-RL (model) & 38.89 \\
  Static Rubric-RL (expert) & 41.83 \\
  EvoRubric (expert) & 44.64 \\
  \midrule
  \textbf{EvoRubric (model)} & \textbf{45.13} \\
  \bottomrule
  \end{tabular}
\end{wraptable}
Expert initialization raises Static Rubric-RL from 38.89 to 41.83 (Table~\ref{tab:expert_eval}), and co-evolution further improves the expert-initialized score to 44.64 (+2.81 points). EvoRubric therefore extends the benefits of expert priors by discovering additional criteria from evolving responses.

Interestingly, model-initialized EvoRubric reaches 45.13, slightly exceeding expert initialization by 0.49 points in this experiment. The results show that co-evolution is effective with both sources of initial criteria and can develop useful supervision from the model's own rubrics.

\subsection{Rubric and Human Evaluation}
\label{sec:rubric_quality_analysis}
\paragraph{Validity and diversity over training.}
Three annotators with medical-rubric annotation experience audited 20 Medical prompts over four training epochs. Table~\ref{tab:human_rubric_summary} reports final-pool validity and effective diversity: the number of human-valid, semantically distinct dimensions beyond the base rubrics, divided by all pool entries. Semantically equivalent criteria count once per prompt in the numerator; the denominator retains all entries.

Effective diversity decreases from 89.19\% to 75.16\% between the first and fourth epoch-end snapshots at 8B, and from 91.62\% to 76.26\% at 14B. The final pools retain 118 and 106 distinct, human-valid dimensions beyond the base rubrics, respectively. The audit characterizes retained rubric coverage, complementing the memory ablation's assessment of its contribution to benchmark performance.

\begin{table}[htbp]
\centering
\caption{Final-pool audit over 20 Medical prompts: counts (\%), with totals $N$ in the header. Effective diversity counts distinct, human-valid dimensions beyond the base rubrics using the same denominators. Percentages are rounded independently; dashes denote unreported results.}
\label{tab:human_rubric_summary}
\label{tab:human_errors}
\begin{tabular}{lrrr}
\toprule
Measure & 8B & 14B & 8B w/o verifier \\
 & $N=157$ & $N=139$ & $N=192$ \\
\midrule
Factual conflict & 4 (2.55) & 3 (2.16) & 30 (15.63) \\
Semantic duplication & 2 (1.27) & 1 (0.72) & 15 (7.81) \\
Style hallucination & 2 (1.27) & 2 (1.44) & 10 (5.21) \\
Unmeasurable & 0 (0.00) & 0 (0.00) & 5 (2.60) \\
Trivial/cliche & 0 (0.00) & 0 (0.00) & 5 (2.60) \\
Length hallucination & 0 (0.00) & 0 (0.00) & 10 (5.21) \\
\midrule
Invalid criteria & 8 (5.10) & 6 (4.32) & 75 (39.06) \\
Valid criteria & 149 (94.90) & 133 (95.68) & 117 (60.94) \\
Effective diversity of the final pool & 118 (75.16) & 106 (76.26) & --- \\
\bottomrule
\end{tabular}
\end{table}

\paragraph{Meta-verification as a validity safeguard.}
Where LOO measures agreement, meta-verification checks the validity of the criteria on which that agreement is based. At 8B, verification reduces factual conflicts from 15.63\% to 2.55\% and unsupported style requirements from 5.21\% to 1.27\%. No length-hallucination criteria were observed with verification, compared with 5.21\% without it. The overall invalid rate falls from 39.06\% to 5.10\%, an 86.96\% relative reduction computed from the underlying counts. These reductions support meta-verification as a safeguard against invalid criteria entering the supervision, complementing the consistency feedback supplied by LOO consensus.

\paragraph{Human response preferences.}
Table~\ref{tab:human_responses} reports a separate blind comparison of final responses. Overall-quality wins exceed losses against both external-generator baselines. The largest advantages occur in conciseness and relevance, whereas completeness favors the baselines more often, factual-correctness outcomes are close, and safety comparisons are mostly ties. The annotation protocol and longitudinal audit details are given in Appendix~\ref{sec:human_audit}.

\begin{table}[htbp]
\centering
\caption{Human response preferences (prompt-level win/tie/loss, \%; 50 Medical prompts). A win favors EvoRubric-8B.}
\label{tab:human_responses}
\begin{tabular}{lcc}
\toprule
Dimension & vs. External & vs. External + GPT-4.1 \\
\midrule
Completeness & 20 / 52 / 28 & 36 / 18 / 46 \\
Factual correctness & 40 / 22 / 38 & 42 / 14 / 44 \\
Conciseness & 54 / 36 / 10 & 68 / 18 / 14 \\
Relevance & 58 / 30 / 12 & 62 / 26 / 12 \\
Safety & 16 / 76 / 8 & 16 / 68 / 16 \\
Overall quality & 42 / 30 / 28 & 48 / 14 / 38 \\
\bottomrule
\end{tabular}
\end{table}

\paragraph{Rubric quality under fixed responses.}
To examine rubric construction with response quality held constant, we compare the two generators on 100 HealthBench queries, each paired with three fixed responses from an unaligned base model. Using the default evaluator from Section~\ref{sec:experimental_setup}, a blind pairwise comparison prefers EvoRubric in 72\% of cases and the same-scale external generator in 28\%, based on professionalism, granularity, and discriminative value (Appendix~\ref{sec:rubric_comparison}). This comparison supports improved rubric construction on identical response groups, complementing the end-to-end benchmark results.

\paragraph{What the learned rubrics capture.}
For the insulin-storage case in Appendix~\ref{sec:appendix_rubric_case_study}, the base rubrics cover general storage conditions. Differences among responses motivate additional checks for formulation-specific durations, handling, device-specific instructions, and source verification. The example illustrates how response variation becomes explicit evaluation criteria for subsequent learning.


\section{Conclusion}
We propose \textbf{EvoRubric}, which constructs rubric-learning supervision for open-ended generation without a verifiable final-answer correctness target. A shared policy discovers evaluation dimensions from its current responses and improves through the supervision those dimensions provide. A frozen copy of the initial policy supplies criterion-validity feedback, while grading, variance filtering, peer consensus, and persistent rubric memory support the joint optimization of the Reasoner and Rubric Generator. Across five benchmarks in Medical, Writing, and Science, EvoRubric outperforms matched static and external evolving-rubric baselines, with macro-average improvements of 3.09 points at 8B and 2.06 points at 14B over the strongest matched baselines across three training seeds. Rubric audits, response comparisons, and expert-initialization experiments provide complementary evidence for this self-evolving approach to open-ended alignment.


\clearpage

\bibliography{iclr2027_conference}
\bibliographystyle{iclr2027_conference}

\clearpage
\appendix
\section{Comparison with Related Methods}
\label{sec:appendix_diff_previous_work}
We compare rubric-based learning paradigms along two dimensions: how evaluation criteria are updated and whether their generation is jointly optimized with the response policy.

\paragraph{Static rubric-based RL.}
Static rubric methods construct evaluation criteria before policy training and use them to score subsequent responses \citep{DBLP:journals/corr/abs-2511-10507, QuRL, RubricRewards}. EvoRubric retains persistent base rubrics and augments them with criteria discovered from current policy responses. A shared policy learns to generate both responses and new criteria, while the Memory Pool carries selected criteria across iterations.

\paragraph{Externally generated evolving rubrics.}
DR-Tulu \citep{DBLP:journals/corr/abs-2511-19399} generates criteria from current responses and existing rubrics, maintains a rubric pool, and filters uninformative criteria. Its reported RL setup uses GPT-4.1 for evolving-rubric generation. EvoRubric learns criterion construction within the response policy, adding validity and peer-agreement feedback. Our GPT-4.1-generator variant follows this external-generation design within the Medical comparison; EvoRubric achieves gains of 2.88/0.83 points on HealthBench/LLMEval-Med (Table~\ref{tab:generator_comparison}). This comparison evaluates the rubric-generation strategy within our training setting. The full DR-Tulu system additionally includes search and citation training.

\paragraph{Joint response and rubric learning.}
RLCER \citep{DBLP:journals/corr/abs-2602-10885} shares parameters across a Reasoner and Rubricator, with role-specific rewards and joint updates. It selects rubrics using their association with final-answer correctness. EvoRubric addresses the absence of that reference: frozen verification assesses criterion validity, discriminative feedback measures response separation, and LOO consensus measures agreement across rubric batches. These provide complementary training signals; consensus itself does not establish correctness.

\paragraph{Joint rubric and judge learning.}
Rubric-ARM \citep{DBLP:journals/corr/abs-2602-01511} alternately trains the rubric generator and judge using preference supervision. EvoRubric updates the Generator and Reasoner while keeping the Grader fixed. The distinction concerns both which components learn and where their supervision comes from.

\section{Training Loop and Joint GRPO Objective}
\label{sec:app_grpo}
Algorithm~\ref{alg:evorubric} summarizes the training loop. Historical criteria used for current-response scoring are retrieved before the memory update; selected additions are carried forward to the next visit to that query.

\begin{algorithm}[htbp]
\caption{EvoRubric training loop}
\label{alg:evorubric}
\begin{algorithmic}[1]
\Require Training queries $\mathcal D$ with fixed base rubrics $\mathcal C_{base}(q)$; initial policy $\pi_{\theta_0}$; frozen Grader; rollout sizes $N,M$; pool capacity $C_{max}$
\State Initialize $\theta\gets\theta_0$ and per-query memory $\mathcal M_q\gets\emptyset$; freeze the Meta-Verifier at $\pi_{\theta_0}$
\While{training continues}
    \State Sample a minibatch of queries from $\mathcal D$
    \For{each query $q$ in the minibatch}
        \State Retrieve $\mathcal C_{hist}\gets\mathcal M_q$ and sample $N$ Reasoner responses $\mathcal O$
        \State Sample $M$ rubric sets conditioned on $q,\mathcal O,\mathcal C_{base}(q),\mathcal C_{hist}$
        \State Meta-verify the proposed criteria to obtain $\{\tilde R_m\}_{m=1}^M$
        \State Grade responses against base, historical, and verified candidate criteria
        \State Compute discriminative and LOO-consensus feedback; identify eligible batches
        \State Compute $R_{gen}$ and $R_{rea}$ using base, historical, and eligible current feedback
        \State Remove saturated entries from $\mathcal M_q$
        \If{at least one candidate batch is eligible}
            \State Admit criteria from the highest-consensus eligible batch after duplicate checks
        \EndIf
        \State Evict oldest entries until $|\mathcal M_q|\leq C_{max}$
        \State Standardize advantages separately for this query's Reasoner and Generator groups
    \EndFor
    \State Update $\theta$ with the combined rollout batch using the clipped GRPO objective
\EndWhile
\end{algorithmic}
\end{algorithm}

For each query, the rollout policy $\pi_{\theta_{old}}$ samples $N$ Reasoner outputs and $M$ Generator outputs. Advantages are standardized separately within each query--role group. Let $\mathcal B$ be the combined rollout batch, $x_b$ the role-specific input, and $y_b$ the sampled output. The implementation averages the clipped objective over response tokens:
\begin{equation}
J(\theta)=\mathbb E\left[\frac{1}{\sum_{b\in\mathcal B}|y_b|}
\sum_{b\in\mathcal B}\sum_{t=1}^{|y_b|}
\min\left(\rho_{b,t}\hat A_b,\operatorname{clip}(\rho_{b,t},1-\epsilon_{low},1+\epsilon_{high})\hat A_b\right)\right],
\end{equation}
where $\rho_{b,t}=\pi_\theta(y_{b,t}\mid x_b,y_{b,<t})/\pi_{\theta_{old}}(y_{b,t}\mid x_b,y_{b,<t})$. Both roles contribute to one shared policy update; their rewards are not standardized together. Adding a numerical stabilizer to the group standard deviation gives zero advantages for constant-reward groups.

\section{Reward Computation}
\label{sec:reward_details}
\paragraph{Base and pool scores.}
For a criterion set $\mathcal C$, let $w_c$ be the signed weight of criterion $c$ and $z_c(o)\in\{0,1\}$ indicate whether $o$ meets it. For a negative-weight criterion, $z_c(o)=1$ means that the penalized behavior occurs. Define
\[
A(o;\mathcal C)=\sum_{c\in\mathcal C}w_c z_c(o),\qquad
P(\mathcal C)=\sum_{c\in\mathcal C}\max(w_c,0),\qquad
N(\mathcal C)=\sum_{c\in\mathcal C}\max(-w_c,0).
\]
All base, historical, and current rubric sets use the same normalized score:
\[
F(o;\mathcal C)=
\begin{cases}
\dfrac{A(o;\mathcal C)}{\max\{P(\mathcal C),N(\mathcal C)\}},
& \max\{P(\mathcal C),N(\mathcal C)\}>0,\\[6pt]
0, & \text{otherwise}.
\end{cases}
\]
The denominator includes all weights in the scored set and is shared across responses. Thus $F\in[-1,1]$, including sets containing only negative weights; empty or zero-weight sets score zero.

Let $\mathcal H$ contain the historical criteria that distinguish the current responses and let $\mathcal E$ index the eligible current rubric batches. Their mean score is
\[
S_{current}(o)=
\begin{cases}
\dfrac{1}{|\mathcal E|}\displaystyle\sum_{m\in\mathcal E}F(o;\tilde R_m),
& |\mathcal E|>0,\\[6pt]
0, & |\mathcal E|=0.
\end{cases}
\]
The two sources of Reasoner supervision are
\[
S_{base}(o)=F(o;\mathcal C_{base}),\qquad
S_{pool}(o)=\tfrac12 F(o;\mathcal H)+\tfrac12 S_{current}(o).
\]
The historical and current terms retain their fixed weights when either is absent. Both $S_{base}$ and $S_{pool}$ lie in $[-1,1]$ and enter $R_{rea}=S_{base}+S_{pool}-p_{len}$. Current supervision averages all eligible batches; only the highest-consensus eligible batch contributes new entries to persistent memory.

\paragraph{Consensus and validity feedback.}
The Meta-Verifier assigns $v(r)=1$ to VALID criteria and $v(r)=0$ to INVALID criteria. For a batch $R_m$ containing $K_m$ proposed criteria before filtering, the Generator's validity reward is
\[
r_{meta}(R_m)=\frac{\sum_{k=1}^{K_m}v(r_{m,k})}{\max\{1,K_m\}}.
\]
This is the fraction of valid criteria among all proposals, with zero assigned to an empty batch. Only valid criteria contribute to subsequent response scoring. Peer consensus uses nonconstant candidate-batch score vectors and is zero when there are no other nonconstant batches or their mean vector is constant. Its positive part is scaled by this validity fraction. The discriminative component measures the fraction of verified criteria that split the responses and is likewise scaled by the validity fraction.

\paragraph{Candidate handling.}
Malformed outputs receive a format reward of $-2$. Empty or fully rejected batches contribute no new criteria. A verified candidate batch is eligible when its response-score vector is nonconstant and its LOO correlation exceeds $0.2$.

\paragraph{Auxiliary regularization.}
For $K$ proposed criteria, the count penalty is $p_{num}=\min\{1,\max\{0,(K-10)/10\}\}$, penalizing outputs beyond the prompted limit of 10 criteria. An additional point-concentration penalty is subtracted from the Generator reward: for $K\geq3$, it is $0.5\max\{0,(f-0.6)/0.4\}$, where $f$ is the fraction of criteria sharing the most frequent absolute weight; it is zero for $K<3$.

\paragraph{Length penalty.}
For a generated output $y$ of token length $\ell(y)$, the overlong penalty is
\[
p_{len}(y)=\alpha\max\!\left(0,\frac{\ell(y)-(L_{max}-L_{buffer})}{L_{buffer}}\right).
\]
Here, $L_{max}=8192$, $L_{buffer}=4096$, and $\alpha=1$. The penalty is zero up to 4096 tokens and grows linearly thereafter. It is subtracted from the reward for both Reasoner and Generator outputs.

\section{Memory Pool and Initialization Details}
\label{sec:memory_details}
Each rollout samples $N=8$ responses and $M=4$ candidate rubric sets per prompt. The Generator receives the fixed base rubrics and all stored historical criteria. For each query, only the eligible batch with the highest LOO consensus can add criteria to the Memory Pool. Saturated historical criteria are removed; candidate entries are checked for exact textual matches against the retained pool before insertion. The pool retains at most 10 entries, evicting the oldest when necessary. Historical criteria that do not distinguish the current responses are excluded from the current historical-pool score. Memory selection retains one batch for future rollouts, whereas current-response supervision aggregates scores from all eligible batches.

Persistent base rubrics are stored separately and remain fixed throughout RL. They are generated once using few-shot prompting with the corresponding frozen pre-RL Qwen3 model. The number of base criteria per prompt ranges from 2 to 8; the proportions for 2, 3, 4, 5, 6, 7, and 8 criteria are 0.03\%, 0.48\%, 4.35\%, 23.17\%, 38.96\%, 13.38\%, and 19.63\%, respectively.

\section{Hyperparameters for Reinforcement Learning}
\label{sec:app_hyperparameters}

In this section, we detail the hyperparameters and configurations used during the reinforcement learning (RL) training phase of our framework. Table~\ref{tab:hyperparameters} provides a comprehensive list of the standard training settings extracted from our experimental setup, including the specific proxy model utilized as the Grader.

\begin{table}[ht]
\centering
\caption{Hyperparameters and configurations for reinforcement learning training.}
\label{tab:hyperparameters}
\begin{tabular}{lc}
\toprule
\textbf{Parameter} & \textbf{Value} \\
\midrule
\multicolumn{2}{l}{\textit{\textbf{Evaluation Configurations}}} \\
Grader & \texttt{GPT-OSS-120B} \citep{DBLP:journals/corr/abs-2508-10925}  \\
\midrule
\multicolumn{2}{l}{\textit{\textbf{GRPO \& Rollout Parameters}}} \\
Rollout Samples per Prompt ($N$) & 8 \\
Rubric Sets per Prompt ($M$) & 4 \\
Evolved Pool Capacity per Prompt & 10 \\
Pool Replacement & FIFO \\
GRPO Clip Bounds ($\epsilon_{low}$, $\epsilon_{high}$) & 0.2, 0.28 \\
KL Loss Coefficient & 0.0 \\
Training Temperature & 0.7 \\
Training Top-$P$ & 0.8 \\
Training Top-$K$ & 20 \\
Max Prompt Length & 12,288 \\
Max Response Length & 8,192 \\
\midrule
\multicolumn{2}{l}{\textit{\textbf{Optimization Parameters}}} \\
Optimizer & AdamW \\
Learning Rate & $1 \times 10^{-6}$ \\
Learning Rate Schedule & Constant \\
Training Batch Size & 64 \\
Mini Batch Size & 64 \\
Micro Batch Size Per GPU & 16 \\
Total Epochs & 4 \\
\midrule
\multicolumn{2}{l}{\textit{\textbf{Reward Shaping Parameters}}} \\
Overlong Buffer Length & 4,096 \\
Overlong Penalty Factor & 1.0 \\
\midrule
\multicolumn{2}{l}{\textit{\textbf{Hardware Setup}}} \\
GPUs & 8 $\times$ NVIDIA H20 \\
\bottomrule
\end{tabular}
\end{table}

\section{Qualitative Case Study: Rubric Evolution}
\label{sec:appendix_rubric_case_study}

Table~\ref{tab:case_study} illustrates rubric evolution for a HealthBench query about insulin storage. The fixed base rubrics cover general temperature, duration, and storage conditions. Differences among the sampled responses motivate additional criteria about formulation-specific duration, handling, device-specific instructions, and source verification. The example illustrates how response-conditioned rubric generation extends the coverage of fixed base criteria.


\begin{longtable}{p{0.96\textwidth}}
\caption{Qualitative case study: EvoRubric dynamically discovers highly discriminative criteria that are missing from generic base rubrics.}
\label{tab:case_study} \\
\toprule
\textbf{User Query:} I am a diabetic on insulin. How do I store it? \\
\midrule

\textbf{Existing Base Rubrics (Generic \& Limited)} \\
\begin{itemize}
    \item Correctly explains that unopened insulin should be refrigerated at $2$--$8^\circ$C, and that opened insulin can generally be stored at room temperature for up to 28 days, depending on the formulation. \textbf{(+10)}
    \item Incorrectly states that insulin can be frozen or exposed to temperatures above $86^\circ$F ($30^\circ$C), which can degrade the medication. \textbf{(-10)}
    \item Mentions avoiding direct sunlight and heat sources during storage. \textbf{(+7)}
\end{itemize} \\
\midrule

\textbf{Reasoner Responses (Selected Snippets Highlighting Diversity)} \\
\textit{Response Snippet A:} ``...Long-acting insulins (e.g., glargine [Lantus], detemir [Levemir], degludec [Tresiba]): Degludec (Tresiba) can be stored for up to 56 days at room temperature according to manufacturer instructions... Cloudy insulins (e.g., NPH) should be gently rolled, not shaken, before use to mix the suspension evenly...'' \\

\textit{Response Snippet B:} ``...If using an insulin pen, store it at room temperature after first use unless the product label explicitly states otherwise. Do not refrigerate opened insulin pens unless instructed by the manufacturer or your healthcare provider...'' \\
\midrule

\textbf{EvoRubric Generation (Ours -- Highly Discriminative \& Nuanced)} \\
\begin{itemize}
    \item \textbf{[Type-Specific Duration]} Provides precise storage durations for different insulin types or formulations (e.g., rapid-acting insulin is often limited to around 28 days, whereas some long-acting formulations such as insulin degludec may allow up to 56 days), rather than relying on a generic 28-day rule. \textbf{(+7)}
    
    \item \textbf{[Handling Nuance]} Explicitly advises against shaking cloudy insulin vials or pens and instead recommends gently rolling them when mixing is required, helping avoid foaming, improper suspension, or dosing inaccuracy. \textbf{(+6)}
    
    \item \textbf{[Device-Specific Rule]} Specifies that in-use insulin pens should generally be stored according to the product label, often at room temperature after first use, while avoiding unnecessary refrigeration unless the manufacturer explicitly permits or recommends it. \textbf{(+6)}
    
    \item \textbf{[Source Verification]} Mentions consulting the manufacturer’s package insert, pharmacist, or healthcare provider for formulation-specific storage instructions, acknowledging that storage durations and conditions vary across insulin products. \textbf{(+5)}
\end{itemize} \\
\bottomrule
\end{longtable}

\section{Controlled Rubric Quality Comparison}\label{sec:rubric_comparison}
For 100 randomly sampled HealthBench queries, an unaligned base model generates three fixed responses per query. The same-scale External Evolving-RL generator and EvoRubric generate rubrics for identical query-response groups. In a blind pairwise comparison, the default evaluator specified in Section~\ref{sec:experimental_setup} selects the more professional, granular, and discriminative rubric set, preferring EvoRubric in 72\% of cases and the external baseline in 28\%. Holding the query and response group fixed isolates differences in the generated rubric sets. A qualitative example appears in Appendix~\ref{sec:appendix_rubric_case_study}.

\section{Effect of External Generator Capacity}
\label{sec:strong_generator_details}
We record generator API usage for the GPT-4.1 variant in Table~\ref{tab:generator_comparison}.
Averaged across the three runs, GPT-4.1 rubric generation consumed 81.20 million input tokens and 2.91 million output tokens per run (84.11 million total per run), costing approximately USD 185.70 per run at the pricing used for the experiment. This estimate covers generator API usage.

\section{Human Evaluation and Rubric Dynamics}
\label{sec:human_audit}
\subsection{Longitudinal rubric audit}
Three annotators with experience in medical-rubric annotation independently audited the same 20 randomly sampled Medical prompts for both model scales. The audit covered generated criteria, Meta-Verifier decisions, and the evolving Memory Pool across all four RL epochs, including the 8B variant without Meta-Verification. Criterion validity was determined by majority vote. Invalid rates are computed by pooling criterion counts across the 20 prompts at each epoch-end Memory Pool snapshot. Pool error and validity percentages use the underlying integer counts and are rounded to two decimal places using round-half-up.

\begin{table}[htbp]
\centering
\caption{Epoch-end Memory Pool audit over 20 Medical prompts. Each entry gives invalid criteria / all audited pool entries (\%). Epoch 4 corresponds to Table~\ref{tab:human_errors}.}
\label{tab:human_validity}
\setlength{\tabcolsep}{3pt}
\begin{tabular}{lcccc}
\toprule
Setting & Epoch 1 & Epoch 2 & Epoch 3 & Epoch 4 \\
\midrule
8B & 2/185 (1.08) & 4/199 (2.01) & 8/177 (4.52) & 8/157 (5.10) \\
14B & 2/167 (1.20) & 3/191 (1.57) & 7/170 (4.12) & 6/139 (4.32) \\
8B w/o verifier & 18/149 (12.08) & 32/132 (24.24) & 52/157 (33.12) & 75/192 (39.06) \\
\bottomrule
\end{tabular}
\end{table}
At epoch 4, the invalid rate is 8/157 with verification and 75/192 without it at 8B, an 86.96\% relative reduction. Final valid-criterion rates are 94.90\% (8B), 95.68\% (14B), and 60.94\% (8B without verification). Residual errors increase over training even with verification, indicating that some invalid criteria persist in the pool. Verifier--human agreement is 78.5\% at 8B and 81.2\% at 14B; it is unavailable without verification. The 14B setting has higher agreement and a lower final invalid rate, although both the policy and verifier scale change in this comparison.

The residual error categories and final validity rates are reported in Table~\ref{tab:human_errors} in the main text.

The audit distinguishes factual conflicts, semantic duplication, unsupported style or length requirements, criteria that cannot be measured, and trivial criteria. Factual conflicts remain the largest residual category with verification. All six categories occur without verification; unmeasurable, trivial, and length-hallucination errors were not observed in the final verified pools of this audit.

\subsection{Effective diversity of the Memory Pool}
We measure effective diversity on the same epoch-end pool snapshots used for the validity audit. For each audited prompt $q\in\mathcal Q$, let $\mathcal M_q^{(e)}$ be its pool at the end of epoch $e$, and let $\mathcal U_q^{(e)}$ contain one representative of each semantically distinct, human-valid evaluation dimension in that pool that is absent from the prompt's fixed base rubrics. We compute
\[
D_e=\frac{\sum_{q\in\mathcal Q}|\mathcal U_q^{(e)}|}
{\sum_{q\in\mathcal Q}|\mathcal M_q^{(e)}|}.
\]
Semantically equivalent eligible entries count once per prompt in the numerator, while all pool entries remain in the denominator. Counts are summed across the 20 prompts before division; criteria from different prompts are not deduplicated together. This within-pool grouping is distinct from the Meta-Verifier's Semantic Duplication check against base rubrics (Appendix~\ref{sec:app_prompts}). Each snapshot measures the dimensions retained at that epoch, including criteria carried over from earlier epochs.

\begin{table}[htbp]
\centering
\caption{Effective diversity of epoch-end Memory Pools. Entries give distinct, human-valid dimensions beyond base / all pool entries (\%), using the same denominators as Table~\ref{tab:human_validity}.}
\label{tab:rubric_diversity}
\setlength{\tabcolsep}{3pt}
\begin{tabular}{lcccc}
\toprule
Setting & Epoch 1 & Epoch 2 & Epoch 3 & Epoch 4 \\
\midrule
8B & 165/185 (89.19) & 172/199 (86.43) & 143/177 (80.79) & 118/157 (75.16) \\
14B & 153/167 (91.62) & 168/191 (87.96) & 139/170 (81.76) & 106/139 (76.26) \\
\bottomrule
\end{tabular}
\end{table}

Effective diversity declines across the four snapshots, with 14B remaining slightly higher at each epoch. At epoch 4, distinct, human-valid dimensions beyond the base rubrics account for 75.16\% and 76.26\% of the 8B and 14B pool sizes. The final pools thus retain substantial non-redundant coverage beyond the fixed base criteria.

\subsection{Blind evaluation of final responses}
The same three annotators evaluated responses on 50 randomly sampled Medical prompts in a blind pairwise comparison. We compare EvoRubric-8B against the same-scale External Evolving-RL baseline and the stronger GPT-4.1-generator variant. The dimensions are completeness (coverage of relevant aspects), factual correctness (accuracy without fabrication), conciseness (absence of unnecessary content), relevance (focus on the question), safety (medically appropriate, non-harmful information), and overall quality. For each prompt and dimension, a judgment supported by at least two annotators is adopted; a three-way split among win, tie, and loss is recorded as a tie.
The dimension-wise results are reported in Table~\ref{tab:human_responses} in the main text.

Overall-quality wins exceed losses against both baselines, corresponding to non-tie win rates of 60.0\% and 55.8\%. The largest advantages are in conciseness and relevance, while completeness favors the baselines more often and factual-correctness outcomes are close. Safety comparisons are mostly ties. These results characterize response-quality trade-offs within the 50-prompt evaluation.


\section{Prompt Templates}
\label{sec:app_prompts}
\begin{longtable}{>{\raggedright\arraybackslash}p{0.96\linewidth}}
\caption{Prompt template for the Rubric Generator.}
\label{tab:prompt_generator} \\
\toprule
\textbf{System Prompt} \\
\midrule
\endfirsthead
\caption[]{Prompt template for the Rubric Generator (Continued)} \\
\toprule
\textbf{System Prompt} \\
\midrule
\endhead
\midrule
\multicolumn{1}{r}{\textit{Continued on next page...}} \\
\endfoot
\bottomrule
\endlastfoot

You are a world-class Rubric Design Specialist for AI Alignment, specializing in Reinforcement Learning from Human Feedback (RLHF). \\

\textbf{\#\#\# Core Mission} \\
Your mission is to analyze a Question, a set of Model Responses, and Existing Rubrics to generate new, highly discriminative evaluation criteria. You must identify critical quality dimensions or specific errors in the text that are \textbf{currently missing} from the existing set of criteria. Your goal is to ensure that any meaningful difference between the responses is captured by a specific rubric. \\

\textbf{\#\#\# Input Data} \\
You will be provided with: \\
1. \textbf{Question}: The original prompt given to the models. \\
2. \textbf{Responses}: A set of text outputs generated by the models. \\
3. \textbf{Existing Rubrics}: The current list of criteria already in use. \\

\textbf{\#\#\# Core Principles: You Must Adhere to These Rules} \\
\textbf{1. Focus on Discriminative Features (The ``Gap'')} \\
\textbullet~ Analyze the content of the responses deeply. Look for significant differences in quality, accuracy, style, or reasoning. \\
\textbullet~ If two responses differ meaningfully (e.g., one is concise and accurate, the other is verbose and vague), but no Existing Rubric captures this difference, you must create a new rubric for it. \\
\textbullet~ Avoid generic criteria that apply to all responses equally. Focus strictly on what makes them \textit{different}. \\

\textbf{2. Radical Non-Redundancy \& Sparsity (CRITICAL)} \\
\textbullet~ \textbf{Check Coverage First}: Before creating a new rubric, rigorously verify if an Existing Rubric already covers the specific quality or error you noticed. \\
\textbullet~ \textbf{Avoid Duplication}: Never duplicate an existing rubric in meaning or scope. Only target specific nuances or dimensions that are completely absent from the current list. \\
\textbullet~ \textbf{Embrace the Empty List}: If the Existing Rubrics are already comprehensive and fully cover the key aspects of the responses, you MUST return an empty list \texttt{[]}. Acknowledging completeness is a sign of expert judgment, not a failure. Do NOT force the creation of mediocre rubrics just to output something. \\

\textbf{3. Style Alignment \& Formatting} \\
\textbullet~ \textbf{Mimic the Style}: Ensure your new criteria match the tone, length, and specificity level of the \textit{Existing Rubrics}. \\
\textbullet~ \textbf{Condition-Based Scoring}: All criteria must be written as conditions that, \textbf{when met}, trigger the point assignment. \\
\textbullet~ \textbf{Positive Rubrics (+ Points)}: Describe a specific excellence found in high-quality responses (e.g., ``Includes a specific Python code example to illustrate the concept'' $\rightarrow$ If present, +2 points). \\
\textbullet~ \textbf{Negative Rubrics (- Points)}: Describe a specific error or flaw found in low-quality responses (e.g., ``Contains hallucinated medical advice'' $\rightarrow$ If present, -5 points). \\
\textbullet~ \textbf{Active Failures Only}: Do NOT write negative rubrics as ``Fails to do X''. Instead, describe the active failure: ``Doing X incorrectly'', ``Providing dangerous workarounds'', or ``Exhibiting Y flaw''. \\

\textbf{4. Actionable \& Objective} \\
\textbullet~ Each criterion must describe an observable, verifiable quality or error in the response text. It should be a clear, unambiguous instruction for both human raters and future reward model training. \\

\textbf{\#\#\# Analysis Workflow} \\
\textbf{1. Mental Evaluation (The Coverage Check)}: \\
\textbullet~ Read the \textbf{Responses} and review the \textbf{Existing Rubrics}. \\
\textbullet~ Mentally apply the existing rubrics to the responses. \\
\textbullet~ Ask: \textit{Is there a significant strength in a good response or a fatal flaw in a bad response that completely slips through the current criteria?} \\

\textbf{2. Gap Identification}: \\
\textbullet~ Identify specific traits (reasoning steps, formatting, safety constraints, tone) that heavily distinguish the responses but are absent from the input rubrics. \\

\textbf{3. Direction \& Formulation}: \\
\textbullet~ Frame the identified quality as a concrete condition to be met. \\

\textbf{4. Prioritization (Quality > Quantity)}: \\
\textbullet~ Generate 0 to 10 total rubrics. \\
\textbullet~ 1 piercingly accurate, highly discriminative rubric is infinitely better than multiple mediocre ones. If no meaningful gap exists, output \texttt{[]}. \\

\textbf{\#\#\# Output Format} \\
Return ONLY a JSON list of objects. Do not include introductory or concluding remarks outside the JSON block. \\
\texttt{[} \\
\texttt{~~\{} \\
\texttt{~~~~"criterion": "<Actionable description of the quality or error>",} \\
\texttt{~~~~"points": <integer, positive for good traits, negative for bad traits, from -10 to 10>} \\
\texttt{~~\},} \\
\texttt{~~...} \\
\texttt{]} \\

\end{longtable}

\begin{longtable}{>{\raggedright\arraybackslash}p{0.96\linewidth}}
\caption{Prompt template for the Meta-Verifier tailored for Medical domain.}
\label{tab:prompt_meta_verifier} \\
\toprule
\textbf{System Prompt} \\
\midrule
\endfirsthead
\caption[]{Prompt template for the Meta-Verifier (Continued)} \\
\toprule
\textbf{System Prompt} \\
\midrule
\endhead
\midrule
\multicolumn{1}{r}{\textit{Continued on next page...}} \\
\endfoot
\bottomrule
\endlastfoot

You are an elite medical grading rubric reviewer. Your task is to evaluate whether a newly proposed grading rubric is valid, practical, and non-redundant compared to existing base rubrics. \\

You must critically evaluate the Target Rubric against the following 6 error categories. If it triggers ANY of these, it is INVALID. \\

\textbf{1. Factual Conflict:} The rubric rewards medically dangerous extrapolation, unverified diagnoses, or sycophancy (e.g., rewarding a response for agreeing with a user's false premise). It contradicts standard medical guidelines or encourages hallucinations. \\
\textbf{2. Unmeasurable:} The rubric demands actions that are impossible for a human rater to verify or clinically impractical for a patient (e.g., ``test the pH of the prescribed ointment at home''). \\
\textbf{3. Semantic Duplication:} The rubric functionally overlaps with an existing base rubric, even if it uses different vocabulary or complex medical jargon to mask the duplication. \\
\textbf{4. Trivial/Clich\'e:} The rubric focuses on generic, superficial advice that applies to almost any medical query (e.g., ``consult a doctor'') unless specifically critical to the prompt context. \\
\textbf{5. Length Hallucination:} The rubric specifically rewards excessively long or verbose answers without adding clinical value. \\
\textbf{6. Style Hallucination:} The rubric enforces a specific formatting style (e.g., ``must use a markdown table'') that is irrelevant to the clinical accuracy of the response. \\

Output your evaluation strictly in the following JSON format. Do not include any other text or markdown blocks outside the JSON: \\
\texttt{\{} \\
\texttt{~~~~"verdict": "VALID" // OR "INVALID: <Category Name>"} \\
\texttt{~~~~// (e.g., "INVALID: Semantic Duplication", "INVALID: Factual Conflict")} \\
\texttt{\}} \\

\end{longtable}

\begin{longtable}{p{0.96\linewidth}}
\caption{Prompt template for the Meta-Verifier tailored for Writing and Science domains.}
\label{tab:meta_verifier_writing_science} \\
\toprule
\textbf{System Prompt} \\
\midrule
\endfirsthead

\multicolumn{1}{c}%
{{\bfseries \tablename\ \thetable{} -- continued from previous page}} \\
\toprule
\textbf{System Prompt} \\
\midrule
\endhead

\midrule
\multicolumn{1}{r}{{Continued on next page...}} \\
\endfoot

\bottomrule
\endlastfoot

You are an elite Meta-Reviewer evaluating grading rubrics for complex, open-ended writing tasks. Your task is to evaluate whether a newly proposed grading rubric is valid, objective, and task-specific compared to existing base rubrics. \newline

You must critically evaluate the Target Rubric against the following 6 error categories. If it triggers ANY of these, it is INVALID. \newline

\begin{itemize}
    \item \textbf{1. Task Constraint Conflict:} The rubric contradicts explicit constraints defined in the original prompt (e.g., word count, specific tone, required elements like quoting experts or mentioning a police officer) OR it hallucinates arbitrary rigid constraints not requested by the user.
    
    \item \textbf{2. Unmeasurable Subjectivity:} The rubric relies on highly subjective aesthetic judgments (e.g., ``beautiful prose'', ``captivating style'') rather than objectively verifiable elements.
    
    \item \textbf{3. Factuality/Safety Conflict:} For prompts dealing with real-world entities, policies, or science, the rubric rewards misinformation, bias, or unsafe content.
    
    \item \textbf{4. Trivial/Generic:} The rubric provides generic writing advice (e.g., ``check spelling'', ``use good grammar'') that applies to any text, completely failing to address the unique semantic goals of this specific prompt.
    
    \item \textbf{5. Length Hallucination:} The rubric specifically rewards or punishes response length UNLESS a strict length constraint was explicitly stated in the user's original prompt.
    
    \item \textbf{6. Semantic Duplication:} The rubric functionally overlaps with an existing base rubric, even if rephrased.
\end{itemize}

Output your evaluation strictly in the following JSON format. Do not include any other text or markdown blocks outside the JSON: \newline

\texttt{\{} \newline
\texttt{~~"verdict": "VALID" // OR "INVALID: <Category Name>"} \newline
\texttt{~~// (e.g., "INVALID: Task Constraint Conflict", "INVALID: Unmeasurable Subjectivity")} \newline
\texttt{\}} \\

\end{longtable}



\end{document}